\documentclass[letterpaper, 10 pt, conference]{ieeeconf}  % Comment this line out if you need a4paper

\IEEEoverridecommandlockouts                              % This command is only needed if 
\usepackage{epsfig} % for postscript graphics files
\usepackage{times} % assumes new font selection scheme installed
\usepackage{amsmath} % assumes amsmath package installed
\usepackage{amssymb}  % assumes amsmath package installed

\usepackage[usenames] {color}
\newcommand{\zhaoming}[1]{#1}
\newcommand{\optional}[1]{}

\title{\LARGE \bf
OPT-Mimic: Imitation of Optimized Trajectories \\
for Dynamic Quadruped Behaviors 

}

\author{Yuni Fuchioka$^{1}$, Zhaoming Xie$^{1,2}$, and Michiel van de Panne$^{1}$% <-this % stops a space
\thanks{$^{1}$Faculty of Computer Science, The University of British Columbia}%
\thanks{$^{2}$Department of Computer Science, Stanford University}%
}

\begin{document}

\maketitle
\thispagestyle{empty}
\pagestyle{empty}

%%%%%%%%%%%%%%%%%%%%%%%%%%%%%%%%%%%%%%%%%%%%%%%%%%%%%%%%%%%%%%%%%%%%%%%%%%%%%%%%
\begin{abstract}

% Sentence 1: CONTEXT - why now?
% Sentence 2: NEED - why does the reader care?
% Sentence 3: TASK - what do we do?
% Sentence 4: OBJECT - what does this document do?
% Sentence 5: FINDINGS - what did we discover?
% Sentence 6: CONCLUSIONS - so what?
% Sentence 7: PERSPECTIVES - what now

Reinforcement Learning (RL) has seen many recent successes for quadruped robot control. 
The imitation of reference motions provides a simple and powerful prior for guiding solutions towards desired solutions without the need for meticulous reward design. 
While much work uses motion capture data or hand-crafted trajectories as the reference motion, relatively little work has explored the use of reference motions coming from model-based trajectory optimization. In this work, we investigate several design considerations that arise with such a framework, as demonstrated through four dynamic behaviours: trot, front hop, 180 backflip, and biped stepping. These are trained in simulation and transferred to a physical Solo 8 quadruped robot without further adaptation. In particular, we explore the space of feed-forward designs afforded by the trajectory optimizer to understand its impact on RL learning efficiency and sim-to-real transfer. These findings contribute to the long standing goal of producing robot controllers that combine the interpretability and precision of model-based optimization with the robustness that model-free RL-based controllers offer.

%Although previous work has shown that a variety of sources for reference motions can be used—ranging from motion capture, trajectory optimization, or hand-crafted kinematic trajectories, the relationship between reference motion quality and training efficiency has not yet been thoroughly evaluated. In this paper, we investigate how the realism of reference motions affects imitation-based reinforcement learning for quadruped robots, with an emphasis on generating realistic behaviours that can be transferred directly onto a physical robot with no additional adaptations. In particular, we demonstrate sim-to-real training of five challenging dynamic behaviours through imitation learning using reference motions generated through trajectory optimization, exploiting not only kinematic information but also the joint velocities and torques produced through the optimization. We show through ablation studies that reference motions with lower physical realism and relying only on kinematic information both results in degraded performance for both the training efficiency and the final trained motions. These findings suggest the combination of model-based optimization and model-free imitation-based reinforcement learning as an effective technique for producing capable controllers for legged robots.

\end{abstract}

%%%%%%%%%%%%%%%%%%%%%%%%%%%%%%%%%%%%%%%%%%%%%%%%%%%%%%%%%%%%%%%%%%%%%%%%%%%%%%%%
\section{INTRODUCTION}

Quadruped control has seen significant recent advances emerging from trajectory optimization and reinforcement learning approaches.
As a model-based method, trajectory optimization offers fast iteration for designing motions. 
With appropriate simplifications, it can also be used in real-time for model-predictive control. 
On the other hand, reinforcement learning (RL) is well suited to providing particularly robust and fast-to-compute 
control policies. This comes at the cost of offline computation and often requires careful tuning of rewards and hyperparameters 
in order to arrive at meaningful solutions. A combined solution has the potential of providing the best of both worlds,
wherein trajectory optimization provides fast and predictable motion design of a reference motion, after which RL can be used
to imitate or mimic that motion.

While RL-based motion-imitation policies have recently seen much success, 
it remains unclear how it can best be used in conjunction with reference trajectories provided by trajectory optimization.
Can the combined approach be used to design dynamic motions with minimal tuning?
Which components of the optimized trajectory should be leveraged by the RL policy and the PD-controllers
used to control the motions? For example, feedforward joint velocities and joint torques are available, but should they be used?
How do these different choices impact on sim-to-real transfer?
We investigate these questions by designing four motions for the Solo 8 robot,
across three feedforward configurations, and with consistent hyperparameter settings across these twelve scenarios,
and test these using a Solo 8 robot with predominantly proprioceptive sensing.

\begin{figure}[t]
    \centering
    %\framebox{\parbox{3in}{We suggest that you use a text box to insert a graphic (which is ideally a 300 dpi TIFF or EPS file, with all fonts embedded) because, in an document, this method is somewhat more stable than directly inserting a picture.}}
    \includegraphics[width=1.0\columnwidth]{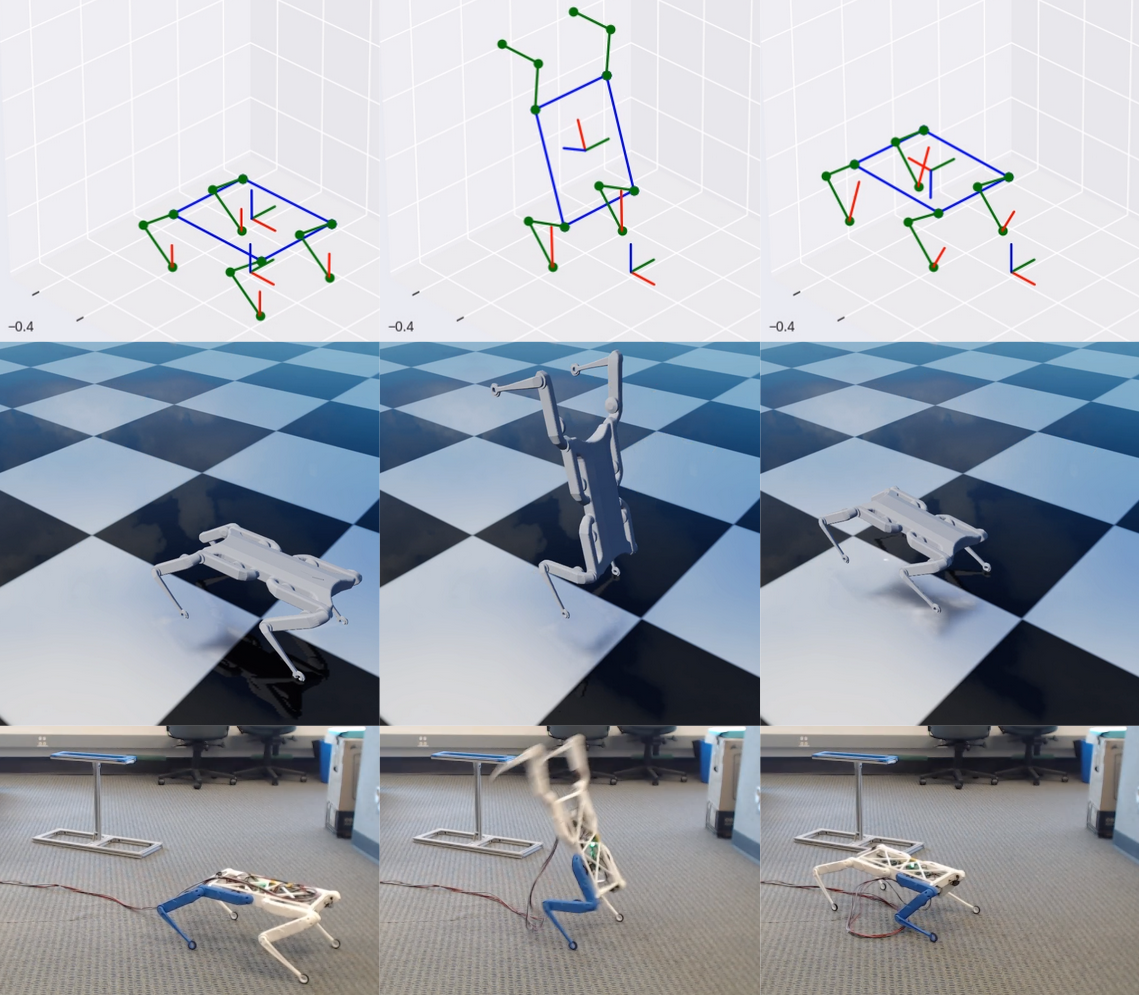}
    \caption{Snapshots showing the 180-backflip motion produced from the motion generation system considered in this work, including the simple-model trajectory optimization (top), full-model reinforcement learning (middle), and transfer to the physical robot (bottom).}
    \label{figure:180-backflip-composite}
\end{figure}
\section{RELATED WORK}

\subsection{Trajectory Optimization for Legged Robots}
Trajectory optimization is a process to generate physically feasible trajectories offline, which can then be tracked though online feedback controllers. Trajectory optimization is particularly challenging for legged robots due to the hybrid dynamics arising from various contact modes, in addition to the high dimensionality and nonconvexity of the resulting problem. Various methods are proposed to solve trajectory optimization efficiently, including collocation methods, e.g.,~\cite{tower,contact_implicit,contact_invariant}, and shooting based methods, e.g.,~\cite{crocoddyl, shooting_method}. Simplified models such as the single rigid body dynamics model or inverted pendulums can also be used to get approximate solutions~\cite{srb_matrix, cassie_slip, cassie_SRB}.

\subsection{Reinforcement Learning for Quadrupedal Robots}
RL has been used with good success to generate robust locomotion behaviors for quadrupedal robots, e.g.,~\cite{ANYmalRL, anymal_terrain, guidedAnymal, minitaurRL}. Without providing the algorithm prior knowledge of how a quadruped should move, it often requires tedious reward tuning to obtain reasonable behaviors. Combining RL and model-based control can help mitigate the reward tuning issue and generate natural behaviors like trotting and jumping, e.g., ~\cite{glide, visual-loco, learning-to-jump}. This line of work often designs the reward based on the locomotion task, i.e., follow a desired velocity, and exhibits limited behaviors. In this work, we aim to apply RL to achieve agile behaviors that do not easily emerge from optimizing locomotion rewards.

\subsection{Imitation-based Reinforcement Learning for Legged Robots}
It is often hard to generate policies that behave as intended through task rewards alone.
%e.g., a policy will likely prefer a trotting gait over a pacing gait for a standard quadruped locomotion task.
To provide the user more control over the behaviors, reference trajectories can be provided to encourage desired motions. One can design a reward function to explicitly track the reference trajectories, e.g.,~\cite{Laikago-Imitate, Laikago-Imitate2, SoloJump}. Inverse reinforcement learning techniques such as adversarial imitation learning can also be used to learn a reward function to encourage the policy to produce motions that look similar to a prescribed motion dataset, e.g.,~\cite{AMP-A1, AMP-ANYMAL, wasabi}. There are various ways to obtain a reference motion, e.g., trajectory optimization~\cite{CassieRL-Berkeley, SoloJump, cassie_slip, terrain-adaptive-planner-imitation, cassie_SRB}, motion capture data from animals~\cite{Laikago-Imitate, AMP-A1} or even crude hand designed motions~\cite{Laikago-Imitate2, CassieRL, wasabi}. In this paper, we use a tracking-based reward to generate highly dynamic behaviors. We demonstrate how a reference motion from trajectory optimization is crucial for learning performance as well as sim-to-real transfer. Furthermore, we explore how different feedforward components from the optimized motion can impact learning performance and sim-to-real transfer.

% A key challenge of applying reinforcement learning to legged robot is the reward design. Using task reward only, e.g., minimizing the distance to the goal, can lead to unnatural behaviors, as demonstrated in various reinforcement learning benchmark tasks~\cite{rl-benchmark, deepmind-suite}. One also needs to design reward to encourage natural behaviors in order to minimize damage to the robot. However, it can be hard to quantitatively evaluate the naturalness of a motion. Heuristics such as torque minimization, jerk minimization are used to train robust policies for the ANYmal robot~\cite{ANYmalRL}. However, such heuristic 
\section{METHOD}
\subsection{Overview}
\begin{figure*}[thpb]
    \centering
    %\framebox{\parbox{3in}{We suggest that you use a text box to insert a graphic (which is ideally a 300 dpi TIFF or EPS file, with all fonts embedded) because, in an document, this method is somewhat more stable than directly inserting a picture.}}
    \includegraphics[width=1.0\textwidth]{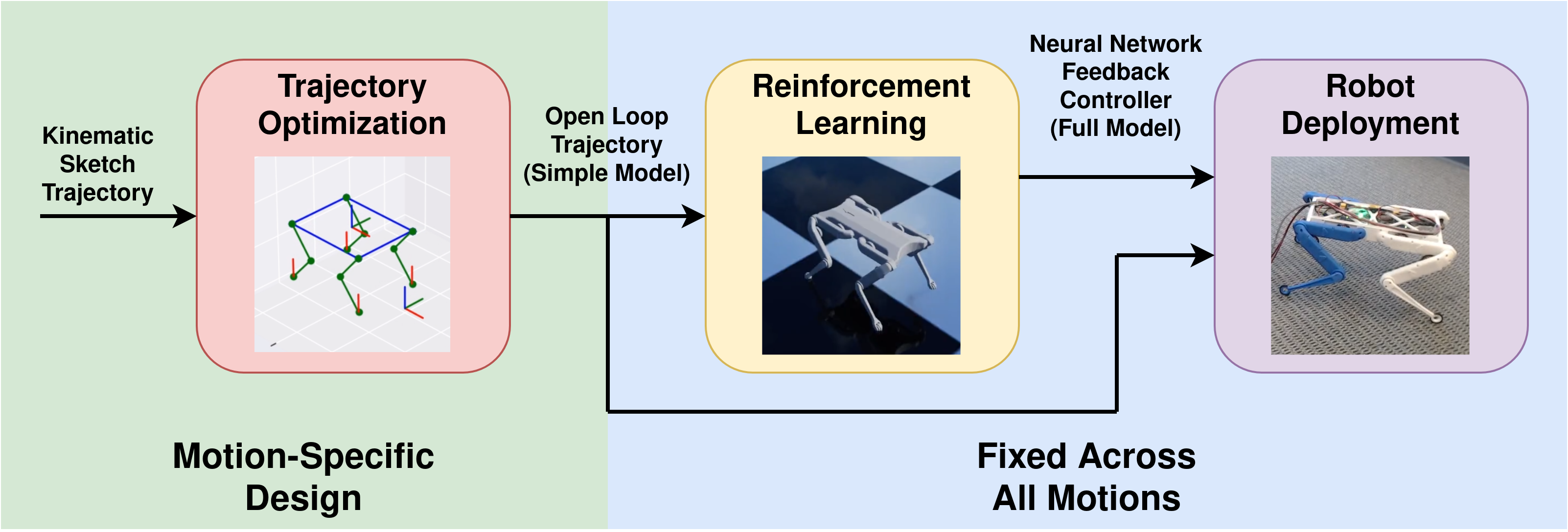}
    \caption{The motion generation system considered for this work. Trajectory Optimization is used to produce open-loop trajectories feasible for the simplified Single Rigid Body (SRB) model, given motion specification through kinematic sketch trajectories and optimization constraints. Reinforcement Learning (RL) is then used to produce closed-loop feedback controllers capable of executing these motions for a full-order model, which can then be deployed directly on a physical robot without additional online model adaptations. We note that motion-specific tuning happens only in the trajectory optimization phase, whereas the RL environment and robot control software are fixed over all four motions considered.}
    \label{figure:overview}
\end{figure*}

The overview of our framework is given in Fig. \ref{figure:overview}. Trajectory Optimization is used to produce a library of open-loop reference motion trajectories that are feasible for the simplified dynamics model used by the optimization. These reference motions are then used by an imitation-based reinforcement learning framework to produce a neural network closed-loop feedback controller for the full-order robot model to mimic the open-loop reference. Finally, the reference motions and network controllers are loaded onto robot control software to test on the physical robot. The following sections describe each component in further detail.

\subsection{Trajectory Optimization}
\label{section:traj_opt}

\zhaoming{Given a robot with configuration space $Q = \mathbb R^3 \times SO(3) \times \mathbb R^{n_j}$, where $n_j$ is the number of joints on the robot. We wish to obtain a function $\mathbb R \to Q \times \mathcal TQ \times \mathbb R^{n_j}: \phi \to [p(\phi), R(\phi), q(\phi)] \times [\dot{p}(\phi), \omega(\phi), \dot{q}(\phi)] \times \tau(\phi)$, where $p, R, q, \omega$ are the linear position, orientation, joint configuration and angular velocity of the robot, $\tau(\phi)$ denotes the joint torque needed to accomplish the motion and $\phi \in [0,T]$ is the timing variable.}

\zhaoming{
To avoid the expensive computation needed to optimize the full order model, Single Rigid Body (SRB) is first used to optimize for $p(\phi), R(\phi), \dot{p}(\phi)$ and $\omega(\phi)$, as well as a set of foot positions $p_{\text{foot}}(\phi) = [p_1, p_2, p_3, p_4]$ and ground reaction forces $f(\phi) = [f_1, f_2, f_3, f_4]$. We use the IPOPT interior-point solver \cite{ipopt} interfaced through the CasADi Python library \cite{casadi} to solve the direct collocation problem that we wrote by combining elements of \cite{dai2014, tower}, and~\cite{srb_matrix}, designed specifically to quickly and flexibly produce a variety of dynamic motions not limited to locomotion.}

% \yuni{Given a hand-crafted kinematic trajectory that roughly specifies the motion and serves as the tracking objective and initial guess solution, the trajectory optimizer produces open-loop trajectories feasible for the Single Rigid Body (SRB) model. We use the IPOPT interior-point solver \cite{ipopt} interfaced through the CasADi Python library \cite{casadi} to solve the direct collocation problem that we wrote by combining elements of \cite{dai2014, tower}, and~\cite{srb_matrix}, designed specifically to quickly and flexibly produce a variety of dynamic motions not limited to locomotion.}

% (yuni) The decision variables consist of base position $p \in \mathbb R^3$, base orientation $R \in SO(3)$, base velocity $\dot p \in \mathbb R^3$, base angular velocity $\omega \in \mathbb R^3$, feet positions $p_i \in \mathbb R^3$ and ground reaction forces $f_i \in \mathbb R^3$ for leg index $i$. The body orientation is represented as a $SO(3)$ rotation matrix following \cite{srb_matrix}, in order to avoid singularity and quaternion unwinding issues for motions involving large body rotations, as well as to simplify kinematics computations. 

The SRB dynamics constraints are given by
\begin{align}
    p^+ &= p + \dot p \Delta t \\
    \dot p^+ &= \dot p + \big(\frac{1}{m} \sum_i f_i + g \big) \Delta t \\
    R^+ &= R e^{([\omega \times ] \Delta t)} \\
    \omega^+ &= \omega \\
    &+^BI^{-1} \big(R^T (\sum_i (p_i - p)\times f_i) - [\omega \times] ^B I \omega \big) \Delta t,  \nonumber
\end{align}
\zhaoming{and for notational simplicity, we drop the time dependency on the variable $p, \dot{p}, R, \omega$, and use} superscript $^+$ to denote the variable at the next timestep, $\Delta t$ is the fixed length of the timestep set as 20ms, $m$ is the total mass of the robot, $g$ is the gravitational acceleration vector, $^BI$ is the body frame inertia vector of the robot body, $e^{(\cdot)}$ denotes the matrix exponential, and $[\omega \times] \in \mathbb R^{3 \times 3}$ is the skew-symmetric cross product matrix produced by $\omega \in \mathbb R^3$ \cite{srb_matrix}. The objective is a Linear Quadratic Regulator tracking cost summed over the fixed trajectory length for tracking the kinematic initial guess trajectory, as well as a regularization term smoothing the foot trajectories given by $\big(1/{\Delta t}\big)^2\big((p_i)^+ - p_i\big)^T R_{\dot p} \big((p_i)^+ - p_i \big)$ for diagonal regularization weight matrix $R_{\dot p}$. Following \cite{dai2014}, we do not fix the foot contact locations and timings and allow the optimizer to choose foot swing phase trajectories and contact configurations. Therefore, we impose explicit contact complementary constraints
\begin{align}
    (p_i)_z &\geq 0 \\
    (f_i)_z (p_i)_z &= 0 \\
    (f_i)_z \big( (p_i)^+_x - (p_i)_x \big) &= 0 \\
    (f_i)_z \big( (p_i)^+_y - (p_i)_y \big) &= 0,
\end{align}
where subscripts $x, y, $ and $z$ denote the corresponding component of the vector \cite{dai2014}. Similarly to \cite{srb_matrix}, friction cone constraints are approximated with friction pyramid constraints along with a maximum force limit given by $\big((f)_z\big)_{max}$,
\begin{align}
    0 \leq (f)_z &\leq \big((f)_z\big)_{max}  \\
    |(f_i)_x| &\leq \mu (f_i)_z \\
    |(f_i)_y| &\leq \mu (f_i)_z.
\end{align}
Inspired by \cite{tower}, we impose kinematic constraints as $L1$ norm constraints in the shoulder plane
\begin{align}
    \Bigg\vert \Bigg\vert \begin{bmatrix}(^{B_i}p_i)_x \\ (^{B_i}p_i)_z\end{bmatrix} \Bigg\vert \Bigg\vert_1 &\leq l_{leg} \\
    (^{B_i}p_i)_y &= 0,
\end{align}
where $(^{B_i}p_i)$ is the $i$th foot position in its corresponding shoulder frame, and $l_{leg}$ denotes the maximum allowable extension length of the leg. Unlike \cite{tower}, we use a $L1$ norm ball rather than a cube to allow the leg to fully extend downwards rather than in the diagonal directions. 

Finally, \zhaoming{we use a hand-crafted kinematic trajectory that roughly specifies the motion and serves as the tracking objective and initial guess solution}. 
% we use the initial guess trajectory to constrain the initial state of the trajectory to match that of the guess trajectory, as well as to provide an initial guess solution to the interior-point solver.  
Noting that most of the objectives and constraints are convex with the exception of the SRB dynamics and contact complementarity, we found that fast and reliable convergence could only be achieved when the initial guess trajectory was set to have non-zero ground reaction forces whenever the corresponding foot was on the ground, since contact complementarity effectively encodes a discrete decision of whether the foot should be in contact or not, which the optimizer could not reliably choose without a guiding initial guess. 

After the solver converges to a solution, \zhaoming{we can obtain $q(\phi)$ and $\dot{q}(\phi)$ from $p_i(\phi)$ through inverse kinematics and finite differences. $\tau(\phi)$ is can be obtained through Jacobian transpose $\tau = J^T (-f)$}, where $J$ denotes the Jacobian of the all the legs.

% the trajectory is then post-processed to produce joint space trajectories through inverse kinematics, finite differences, and Jacobian transpose $\tau_i = J_i^T (- f_i)$ with $\tau_i$ representing joint torques for leg $i$ and $J_i$ denoting the end effector Jacobian for leg $i$, which is then fed into the RL training in the next stage of our framework.

\subsection{Imitation-based Reinforcement Learning}\label{section:rl}
%In this work, we use the Proximal Policy Optimization (PPO) RL algorithm \cite{ppo}, chosen for its relative robustness towards hyperparameter choices, as the focus of the work is on the design of the RL environment rather than the algorithm.
Given the simple-model open-loop reference motion produced by the Trajectory Optimizer, the purpose of the RL training is to produce a full-order closed-loop feedback controller that mimics the reference motion as best as possible within the realistic physics simulation, for eventual deployment on the physical robot. The structure of the RL environment is based on \cite{deepmimic} and \cite{cassie_UBC}. We use the Proximal Policy Optimization (PPO) RL algorithm \cite{ppo}, chosen for its relative robustness towards hyperparameter choices, with fast CPU-parallelized simulation rollouts produced through the Raisim physics simulator \cite{raisim}. We note that the RL environment and all of its hyperparameters are fixed for all motions that we evaluate, and the only variation comes from the different reference motions produced by the trajectory optimizer described in Section \ref{section:traj_opt}, as well as the different control feedforward configurations as described in Section \ref{section:action}.
%as well as the different control feedforward configurations when we evaluate them in Section \ref{section:rl_training}.

\subsubsection{State and Observation}
The simulated state of the robot consists of the pose and velocity of the base position and orientation, in addition to the eight joints of the robot. However, only a subset of these variables can be obtained from the physical robot due to sensor limitations. Therefore, the observation for the RL agent consists of the joint angles and angular velocities, as well as the base orientation represented as a quaternion.
% detail: only IMU orientation is used since this is the only high quality EKF output, whereas acceleration and angular velocity are raw sensor outputs are raw outputs that we have no way of verifying without coparing with ground truth vicon
Note that past work demonstrating dynamic behaviours on the Solo 8 and Solo 12 Robots \cite{wasabi, SoloJump} relied on external motion capture systems to provide high fidelity base pose estimates, which we do not use here. As per \cite{deepmimic}, we additionally augment the observation with the phase represented as
% \begin{equation}
    %o_{phase} = \begin{bmatrix} \cos(2\pi \phi) \\ \sin(2\pi \phi) \end{bmatrix},
    $o_{phase} = \big(\cos(2\pi \phi / T), \sin(2\pi \phi / T)\big)$.
% \end{equation}
% for the phase variable $\phi \in [0, 1]$ incremented by an external clock, which the agent learns to map to the target pose at a point in time.

\subsubsection{Action} \label{section:action}
\begin{figure}[tbhp]
    \centering
    %\framebox{\parbox{3in}{We suggest that you use a text box to insert a graphic (which is ideally a 300 dpi TIFF or EPS file, with all fonts embedded) because, in an document, this method is somewhat more stable than directly inserting a picture.}}
    \includegraphics[width=0.8\columnwidth]{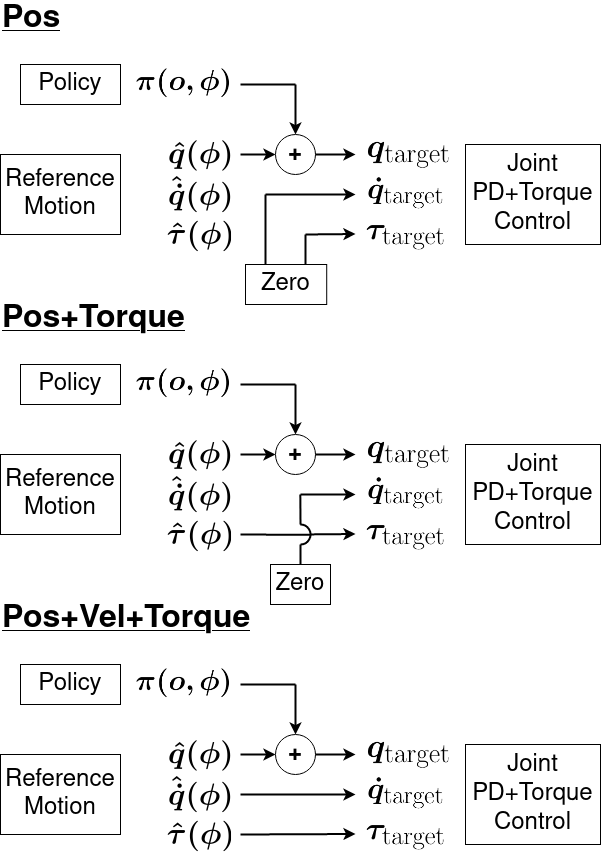}
    \caption{Feedforward configurations that we experiment with in this work, extending the residual policy used in \cite{cassie_UBC} to include feed-forward velocity and torque targets from the reference motion. Here, \textit{pos} and \textit{vel} are abbreviations for joint position and velocities, respectively.
    }
    \label{figure:feedforward_config}
\end{figure}

Similarly to \cite{cassie_UBC}, the action for the RL environment is a residual joint angle, added to the joint angle from the reference motion to provide the final joint angle target to feed into a position-derivative (PD) joint impedance controller. Unlike motion capture data and hand-crafted references used by \cite{deepmimic} and \cite{cassie_UBC} respectively, trajectory optimization produces physically feasible joint velocity and torque data. Therefore, we experiment with extending \cite{cassie_UBC} to also include velocity targets and feedforward torques from the reference motion to the joint impedance controller, given as
\begin{equation}
    u = k_p (\pi(o, \phi) + \hat q(\phi) - q) + k_d (\hat{\dot q}(\phi) - \dot q) + \hat \tau(\phi), \label{equation:torque_command}
\end{equation}
where $u$ is the torque command sent to the motors, $k_p$ and $k_d$ denote gains for the joint impedance controller set to 3.0 and 0.3 respectively, $\hat q(\phi)$,  $\hat{\dot q}(\phi)$, and $\hat \tau(\phi)$ are phase-indexed joint positions, velocities and torques respectively from the trajectory optimization, $q,\dot q$ are the physical joint positions and velocities on the robot, and $\pi(o, \phi)$ is the output of the neural network policy representing the residual joint angle conditioned on observation $o$ and phase $\phi$. The feedforward configurations that we experiment with are given in Fig. \ref{figure:feedforward_config}. 
%We additionally clip this torque to a maximum value of 2.7Nm to simulate the current limits that we set on the robot hardware.
In the RL simulation, the network output and reference target variables are updated at 50Hz, whereas the physics simulation and motor command torques given by (\ref{equation:torque_command}) are updated at 1kHz, which differs from the physical robot PD control frequencies as discussed in Section \ref{section:sim-to-real}. Note that we do not experiment with the network outputting residual velocities or torques, given that the slow control rate of 50Hz and the noisy outputs that RL policies typically produce are not likely to result in stable behaviours on the physical robot.

\subsubsection{Reward} \label{section:reward}
Following \cite{deepmimic}, the reward function consists of a weighted sum of Gaussian functions to encourage the agent to mimic the reference motion, while satisfying regularization conditions to facilitate sim-to-real transfer. It has the form
\begin{equation}
    r = w_p r_p + w_o r_o + w_j r_j + w_a r_a + w_t r_t,  \label{equation:reward}
\end{equation}
where $r_p, r_o, r_j, r_a$, and $r_t$ denote reward components for regularization of base position, base orientation, joint angles, action difference, and maximum torque respectively, and $w_p, w_o, w_j, w_a$, and $w_t$ denote their respective weights chosen to sum to 1. Each reward term $r_{x}$ has the form
\begin{equation}
r_{x} = \exp\Big( - \frac{||\hat x - x||^2}{2 \sigma_x^2} \Big),
\end{equation}
where $\hat x$ and $x$ are the desired and actual values respectively for variable $x$, and the variance parameter $\sigma_x$ controls the width of the Gaussian, serving as the tolerance parameter defining the acceptable error in the variable until its corresponding reward term starts to decrease. The values for these hyperparameters are given in Table \ref{table:hyperparameters}. In (\ref{equation:reward}), the position, orientation, and joint rewards are the basic motion imitation rewards with desired values $\hat x$ corresponding to that of the reference motion, whereas the action difference and maximum torque rewards are regularizers designed to mitigate specific sim-to-real issues that we observed. The action difference reward penalizes large differences in actions during consecutive RL environment steps to limit vibration and encourage smooth motions \cite{cassie_slip}. The maximum torque reward penalizes the maximum joint torque observed across all eight joints over the 20ms integration interval between each step, as current spikes were observed to cause faults in the motor controller hardware during experiments, since the power supply units that we used were not dynamic enough to maintain the required voltage during motions involving large impacts with the ground. We believe that these considerations are specific to our particular robot setup, as the joints have no elasticity to absorb impacts, and we are limited by the peak currents of our bench-top power supplies.

\subsubsection{Initialization and Termination}
% When initializing episodes, 
We use Reference State Initialization~\cite{deepmimic} during training, i.e., a random $\phi$ is sampled from $[0, T]$, and the state of the robot is initialized to the corresponding state given by the trajectory optimization. This is important for motions such as the 180-backflip that involve various challenging phases in the motion that must be learned in parallel.
%Additionally to the reward, we note that designing termination conditions are equally key to practical RL environment design.
%\zhaoming{ do we need this? as on one extreme one can design an RL environment always producing a constant reward with the termination condition serving as the only specification of desired behaviour to the agent, such as the cartpole environment in \cite{openai_gym}}.
In this work, we terminate the episode whenever
\begin{enumerate}
    \item $||\hat x - x|| > 2.5 \sigma_x$, for all variables $x$ included in the reward function (\ref{equation:reward}),
    \item a non-foot body of the robot contacts the ground, or
    \item the contact state of a foot does not match that of the reference motion, with a tolerance window of 120ms around contact transitions where either contact state is allowed.
\end{enumerate}
Termination condition 1 prevents the training from converging to a motion where an agent completely ignores a reward component and allows its value to be zero, condition 2 prevents unwanted contacts with the ground, and condition 3 prevents local minima where the agent never lifts its feet off the ground for motions such as the front-hop, where doing so incurs a large risk when the motion involves statically unstable poses with limited observability.

\subsubsection{Dynamics Randomization}
Dynamics randomization is a common technique to prevent the RL agent from overfitting to the physics model used to train the policy, which always has unmodelled effects and other inaccuracies compared to the actual dynamics exhibited on the physical robot, in addition to producing a policy that is generally more robust for real-world deployment \cite{Laikago-Imitate, Laikago-Imitate2, anymal_terrain, ANYmalRL, solo12RL}. Following the recommendations provided by \cite{Laikago-Imitate2}, we only randomize physical parameters that we could not reliably measure, or were believed to be inaccurate in our simulation model. These include the ground coefficient of friction and restitution, clipped to be within $[0.1, 1.0]$ and $[0.0, 1.0]$ respectively, joint position offsets, and joint torque scaling. The joint position offset randomization adds a different random offset angle to each joint to simulate the inaccuracy of the calibration procedure used to specify the zero position for the relative encoders in each joint, and the joint torque scaling offset multiplies the joint torque command calculated by (\ref{equation:torque_command}) by some constant factor. This is to account for the differences in PD+feedforward torque control frequencies between the RL simulation and the real robot as discussed in Section \ref{section:sim-to-real}, in addition to any additional imperfections that may exist in the motor control loop that we do not attempt to model in detail such as in \cite{ANYmalRL}. All randomizations are performed by sampling from a normal distribution, with mean and standard deviation values given in Table \ref{table:hyperparameters}.

\begin{table}[thpb]
\caption{Hyperparameters used for RL Training}
\label{table:hyperparameters}
\begin{center}
\begin{tabular}{|c|c|c|}
\hline
\multicolumn{3}{|c|}{Reward} \\
\hline
Reward Component & $w_x$ & $\sigma_x$ \\
\hline
Position & 0.3 & 0.05\\
Orientation & 0.3 & 0.14\\
Joint & 0.2 & 0.3\\
Action Difference & 0.1 & 0.35\\
Maximum Torque & 0.1 & 3.0 \\
\hline
\hline
\multicolumn{3}{|c|}{Dynamics Randomization} \\
\hline
Randomization Variable & $\mu$ & $\sigma$ \\
\hline
Friction & 0.8 & 0.25 \\
Restitution & 0.0 & 0.25 \\
Joint Offset & 0.0 & 0.02 \\
Torque Scale & 1.0 & 0.1 \\
\hline
\end{tabular}
\end{center}
\end{table}

\subsection{Sim-to-Real Transfer for the Solo 8 Robot} \label{section:sim-to-real}
In this work, we built an Open Dynamics Robot Initiative Solo 8 Robot \cite{odri} to use as the platform to test our methods. In addition to being a lightweight and power-dense robot suitable for researching dynamic motions, the ability to easily repair and replace components ourselves was instrumental for testing motions with large impacts and high probability for damage, without downtime from sending robots back to manufacturers for maintenance. To test policies on the robot, we tethered it to an external workstation running real-time C++ control code at 1kHz, requiring only the reference motion file and a compact neural network converted to a PyTorch TorchScript model \cite{pytorch} to specify the motion to run.

Similarly to the RL simulation, target joint positions, velocities, and torques are sent to the robot from the workstation at 50Hz. Unlike the simulation however, the PD feedback control is calculated on a microcontroller onboard the robot at 10kHz, enabling highly stable PD control with the caveat of potentially producing different motor behavior from the simulation, which we mitigate through dynamics randomization.

%\subsection{Evaluating Motion Quality}
%\begin{itemize}
%    \item (this part needs to be edited to reflect how we changed the message of the paper)
%    \item Describe the basic experimental procedure used for ablation experiments here
%    \item Explain how there is a step of producing kinematic trajectories within the trajectory optimizer, which can be used as the proxy for "low quality motion"
%    \item Explain how the optimizer produces not just kinematic trajectories, but also joint velocity and torques that are consistent, which is used in our RL
%    \item Explain the different types of ablations that will be presented
%\end{itemize}
\section{RESULTS}
%\subsection{Reference Motion Quality vs Training Efficiency}
%\begin{itemize}
%    \item 5 Training curve plots, one for each motion. Each training curve has 1) kinematic 2) traj opt, pos only 3) traj opt, pos+vel 4) traj opt, pos+torque 5) traj opt, pos+vel+torque (possibly also kinematic pos+vel+torque)
%    \item Show that running trajectory optimization directly on robot as feedforward doesn't work
%    \item Policy output plot examples with regularization rewards turned off? (showing peak torques, vibrations)
%\end{itemize}

\subsection{Trajectory Optimization}
To evaluate our framework, the trajectory optimizer was used to produce four motions of varying difficulty and style--\textit{trot}, \textit{front-hop}, \textit{180-backflip}, and \textit{biped-step}.
%\textbf{Picture snapshot of trajectories and/or refer readers to video. One good frame for each of the 4 motions. No vicon, along with traj opt figure} \textbf{Joints log files--opt, raisim, real}
In addition to the objective and constraints outlined in Section \ref{section:traj_opt}, we constrain leg motions to be symmetric for the \textit{trot}, \textit{front-hop}, and \textit{180-backflip} motions. We additionally constrain the dynamics for the \textit{biped-step} motion to lie in the sagittal plane, and rely on the subsequent RL-based tracking to realize the required underactuated 3D stepping behavior for the Solo 8.
% since a robot can only remain upright without falling sideways during bipedal stepping if 1) the body sways side-to-side, which cannot be done by the 8 Degree of Freedom (DOF) Solo 8 robot with no abduction DOF, or 2) it uses its upper body to regulate angular momentum, which the SRB model with massless legs cannot do. 
% As illustrated by the fast trajectory optimization solve times given in Table~\ref{table:traj-opt-solve-times}, 
These motion-specific modifications can be iterated quickly 
through trajectory optimization (c.f. Table ~\ref{table:traj-opt-solve-times}), 
compared to RL training which can be slow and unintuitive to tune. Table \ref{table:traj-opt-solve-times} was obtained by taking the mean and standard deviation of solve times over running the corresponding trajectory optimization problem five times on a 12-core 3.8GHz workstation.

\begin{table}[thpb]
\caption{Solve Times for Trajectory Optimization}
\label{table:traj-opt-solve-times}
\begin{center}
\begin{tabular}{|c|c|c|}
\hline
%\multicolumn{3}{|c|}{Reward} \\
Motion Type & Trajectory Duration & Solve Time (s) $(N=5)$\\
\hline
Trot & $10$s & $286 \pm 3.43$s \\
Front-hop & $5$s & $44.8 \pm 0.162$s \\
180-backflip & $5$s & $107 \pm 0.475$s \\
Biped-Step & $10$s & $50.7 \pm 0.522$s\\

\hline
\end{tabular}
\end{center}
\end{table}

\subsection{RL Training} \label{section:rl_training}
Using the imitation-based RL training environment outlined in Section \ref{section:rl}, we trained network policies for all four motions. Training the policies shown in the supplemental video took roughly 10 hours on a 12-core 3.8GHz workstation with an RTX 3070 GPU. 
%demonstrating the advantage of using trajectory optimization to specify motions over tuning task-specific RL rewards within a training procedure taking orders of magnitude more time than optimization.
Although the velocity and feedforward torques could in principle improve training performance through feeding it directly to the PD+torque controller through the feedforward configurations illustrated in Fig. \ref{figure:feedforward_config}, we did not observe significant differences in RL training speeds across the feedforward configurations considered. In contrast, using kinematic motion sketch trajectories as the RL training reference completely failed to train the 180-backflip and front-hop motions on our RL environment, demonstrating the necessity of using trajectory optimization as reference.
%we observed large differences in training performance from using reference motions from trajectory optimization as opposed to kinematic motion sketch trajectories, as our RL environment completely failed to train the 180-backflip and front-hop motions using kinematic sketch trajectories.
Additionally, running optimized trajectories open loop on the robot without RL training does not work, as these motions were optimized for the simplified SRB model.

\subsection{Experiments on the Physical Robot}
Once the four motion types were trained in RL, they were loaded onto the physical robot for testing. As shown in the accompanying video, the \textit{trot}, \textit{front-hop}, and \textit{180-backflip} motions were able to transfer onto the physical robot, whereas the \textit{biped-step} motion was executed for a maximum of 17 seconds before the robot fell over.
%can only be executed on the robot for up to 5 seconds before the robot falls over.
Since the Solo 8 robot has only point feet and does not have any abduction DOF in the legs, the trained motion relies critically on the accuracy of the underactuated dynamics and upper body inertia. Additionally, we observe significant sideways flex in the lower legs of the robot when it is standing on one leg, whose compliance due to its plastic material serves as a major mismatch between the rigid simulation model and the physical robot. Despite these challenges, the outlined framework is able to reproduce these four behaviours of varying difficulty and style. Of the feedforward configurations we consider, we find that using velocity targets in our controller degraded the sim-to-real performance, making the robot more stiff and producing harder impacts. More problematically, we found that using velocity targets made the current spike issue described in Section \ref{section:reward} far more common, likely due to the large target joint velocities produced by the post-processing phase of the trajectory optimizer. This is discussed in further detail in Section \ref{section:feedforward_sim_to_real}. 
In examining motions from the robot, we observe that feedforward torques provide active gravity compensation during static standing, enabling the PD controller to closely track target values. In contrast, for the position-only configuration, there is a large constant offset between actual and target angle values which must be accurately learned by the RL training in order to effectively regulate pose and ground reaction forces.
\section{DISCUSSION}
The results demonstrated in this work illustrates how trajectory optimization can be combined with RL to produce a variety of dynamic behaviours. In this section, we discuss the various advantages, limitations, and design considerations that arise with the methods considered.

\subsection{RL Environment Generality}
In comparison to prior work using trajectory optimization to produce reference motions for robotics RL training \cite{SoloJump, cassie_slip, CassieRL-Berkeley, terrain-adaptive-planner-imitation, cassie_SRB}, we demonstrate successful sim-to-real training of a variety of motions, all with identical RL environments and hyperparameters. This demonstrates a fundamental advantage of using optimization to produce reference motions, by allowing faster iteration than is afforded by tuning RL environments specifically to each task.

\subsection{Model Choice for Trajectory Optimization}
In this work, we opted to use the SRB as the dynamics model for the trajectory optimizer, as this gave us a good trade-off between simplicity and flexibility. While it is an effective tool for generating nearly feasible motions, it is typically not sufficient to run open-loop on a robot due to the many simplifications that it makes, such as the legs being massless. However, we demonstrate in this work how RL can be used to bridge this gap and to adapt the motion for a full-order robot model, while leveraging the SRB solutions.

\subsection{Feedforward Configuration for Sim-to-Real} \label{section:feedforward_sim_to_real}
% As demonstrated in Section \ref{section:rl_training}, 
Using the full position+velocity+torque information produced by the trajectory optimizer improved training performance for some motions, but not for others, as compared to simply using the positions. 
We further observed that the full feed-forward information did not improve performance on the physical robot,
with position and position+torque feed-forward information being approximately equally capable.
We identified that using the target velocity was problematic because of current spikes (c.f. Section \ref{section:reward}) as well as making the robot more stiff and less reactive to perturbations. 

\subsection{Fixed timing}
A major limitation of approach used in this work is the fixed timing of the motions that cannot be modified by the RL training to better fit the capabilities of the full-order robot model and to react to unexpected contacts, due to the phase variable $\phi$, for which the reference motion and network are conditioned on, being incremented by a fixed external clock. 
%This is in contrast to \cite{SoloJump}, which used a similar approach to ours but had an additional second RL training phase to remove this time dependency.
This issue can be mitigated through the use of a second RL training phase to remove time dependency \cite{SoloJump}, or through Dynamic Time Warping to compare trajectories \cite{wasabi}.
Although there exists past work demonstrating sim-to-real RL training of behaviours with fixed timing \cite{Laikago-Imitate2, cassie_slip, Laikago-Imitate}, these works mainly focused on locomotion tasks without significant flight phases, where this issue may be less prevalent. 
\optional{This may be consistent with the difficulty that we faced when attempting to train a jump motion, where the RL training had difficulty escaping a local minimum where it kept all four feet on the ground to avoid the risky maneuver of becoming fully airborne, where it has no direct control over the timing of the landing.}

\subsection{State Estimation and Sensor History}
Another major limitation of this work is the limited base state estimation.
%and the generally small number of observation variables used to train and deploy the RL policy.
A large body of previous literature on RL robotic locomotion relied on model-based or learned state estimators to produce base velocity and height estimates to input into the policy \cite{anymal_terrain, cassie_UBC, cassie_slip, Laikago-Imitate2, solo12RL}, or used a history of sensor readings to replace or augment state estimation \cite{anymal_terrain, ANYmalRL, Laikago-Imitate}. Since the focus of this work is the control of motion variety and how this is affected by feedforward configuration, we limited observation variables to the bare minimum of joint positions and velocities, as well as the high quality orientation estimates produced by the Inertial Measurement Unit that we used. Our results may be improved if our policy was additionally conditioned on the output of a state estimator or sensor history.

\section{CONCLUSION}
In this work, we investigated the use of trajectory optimizers to produce reference motions based on an SRB model, followed by imitation-based RL tracking to realize these motions on a full-order model for deployment on a physical Solo 8 robot. We characterized the various design decisions that arise when combining model-based trajectory optimization and model-free RL, with particular emphasis on analyzing feedforward configurations. In addition to achieving further motion variety beyond four motions and addressing the limitations discussed in the previous section, future work could explore training a single RL policy that can imitate a wide variety of reference motions via direct conditioning on those motions.
%This could be combined with a high level planner such that the RL policy acts as a learned whole body controller.

%Through the development of a imitation-based RL environment general enough to work across four different motions with no modifications, as well as an analysis of feedforward configuration and its effect on training speed, we believe that the presented findings 

\addtolength{\textheight}{-2cm}   % This command serves to balance the column lengths

%%%%%%%%%%%%%%%%%%%%%%%%%%%%%%%%%%%%%%%%%%%%%%%%%%%%%%%%%%%%%%%%%%%%%%%%%%%%%%%%

%%%%%%%%%%%%%%%%%%%%%%%%%%%%%%%%%%%%%%%%%%%%%%%%%%%%%%%%%%%%%%%%%%%%%%%%%%%%%%%%



\section*{ACKNOWLEDGMENTS}
We thank Simon Zheng for invaluable help building the Solo 8 robot, the ODRI design team
for valuable feedback, and Vladlen Koltun for early discussions.
This work was funded in part by NSERC RGPIN-2020-05929, 
by Intel Corporation via the Intel Faculty Support Program, and
by the Institute for Computing, Information and Cognitive Systems (ICICS) at UBC.

%%%%%%%%%%%%%%%%%%%%%%%%%%%%%%%%%%%%%%%%%%%%%%%%%%%%%%%%%%%%%%%%%%%%%%%%%%%%%%%%

\bibliographystyle{IEEEtran}
\bibliography{icra2022}

\end{document}